\documentclass[lettersize,journal]{IEEEtran}
\usepackage{amsmath,amsfonts}
\usepackage{algorithmic}
\usepackage{algorithm}
\usepackage{array}
\usepackage[caption=false,font=normalsize,labelfont=sf,textfont=sf]{subfig}
\usepackage{textcomp}
\usepackage{stfloats}
\usepackage{url}
\usepackage{verbatim}
\usepackage{graphicx}
\usepackage{cite}
\begin{document}

\title{Artificial Intelligence Based Navigation in Quasi Structured Environment}

\author{Hariram Sampath Kumar, \IEEEmembership{Member, IEEE,} Archana Singh, Manish Kumar Ojha

\thanks{Hariram Sampath Kumar is with the Department of Artificial Intelligence, Amity University, Noida, Uttar Pradesh - 201313, India (e-mail: harisb4160176@gmail.com).}
\thanks{Archana Singh is with the Department of Artificial Intelligence, Amity University, Noida, Uttar Pradesh - 201313, India (e-mail: asingh27@amity.edu).}
\thanks{Manish Kumar Ojha is with the Department of Artificial Intelligence, Amity University, Noida, Uttar Pradesh - 201313, India (e-mail: mkojha@amity.edu).}}

\maketitle

\begin{abstract}
The proper planning of different types of public transportation such as metro, highway, waterways, and so on, can increase the efficiency, reduce the congestion and improve the safety of the country. There are certain challenges associated with route planning, such as high cost of implementation, need for adequate resource \& infrastructure and resistance to change. The goal of this research is to examine the working, applications, complexity factors, advantages \& disadvantages of Floyed-Warshall,
Bellman-Ford, Johnson, Ant Colony Optimization (ACO), Particle Swarm Optimization (PSO), \& Grey Wolf Optimizer (GWO), to find the best choice for the above application. In this paper, comparative analysis of above-mentioned algorithms is presented. The Floyd- Warshall method and ACO algorithm are chosen based on the comparisons. Also, a combination of modified
Floyd-Warshall with ACO algorithm is proposed. The proposed algorithm showed better results with less time complexity, when applied on randomly structured points within a boundary called quasi-structured points. In addition, this paper also discusses the future works of integrating Floyd-Warshall with ACO to develop a real-time model for overcoming above mentioned-challenges during transportation route planning. 
\end{abstract}


\section{Introduction}
Based on the literature survey, a research gap was found for application of Artificial Intelligence (AI) in Transportation Route Planning. In this research, a detailed study is done to find the optimal algorithm for the application and also, developed a new modified algorithm for establishing a research pathway to develop an AI model in different stages for the mentioned application.

The proposed algorithm focuses on initial stage of implementation for transportation route planning. The current implementation stage is split up as follows: 

\begin{enumerate}
    \item Based on comparative analysis choose the best algorithm suitable for the application.
    \item Develop a modified Floyd-Warshall with ACO algorithm and apply on quasi-structured points.
    \item Recorded the results and compared with general Floyd-Warshall algorithm.
\end{enumerate}

Further in this section, we will see about different path-generation algorithms \& nature-inspired optimization algorithms, and comparative analysis of ACO, PSO, GWO, Bellman-Ford, Johnson, Floyd-Warshall. In the upcoming sections, a detailed review of various research \& survey papers has been included. In section 3, a detailed methodology of the research and process of building modified Floyd- Warshall with ACO algorithm has been discussed. In section 4, the results acquired using the proposed algorithm and the classic algorithm is presented and the betterment is discussed. Followed by in the final section, a short brief about the objectives achieved \& the future works has been discussed.

\subsection{Overview of Path Generation Algorithms}

Path generation algorithms are computational methods used to find all the paths in a network between two or more points, taking into account constraints such as node weights and edge weights. This algorithm plays a role in various disciplines, such as transportation, logistics, \& network routing, among others \cite{sapundzhi}.

The Bellman-Ford methodology is a known method for figuring out the shortest route between the origin node and an adjacent node in a network with negative edge weights. It operates by iterating over all graph nodes and relaxing the edges, thereby updating each corresponding node with computed shortest possible route until convergence. This algorithm can manage networks with adverse edge weights, but as a time complexity of O($|V|$$|E|$), making it less effective for vast graphs \cite{zhao}.

The Floyd-Warshall’s algorithm is another classic algorithm to determine the all-possible shortest path between any numebr of points in a graph with positive or negative edge weights. It operates by constructing a table of shortest path estimates between all pairs of graphical points by taking intermediate nodes into account. This algorithm's time issues are O($|V|$3), making it inefficient for vast graphs \cite{risald}, \cite{zhao}.

The Johnson algorithm is a more recent algorithm for locating the fastest route between every pair of a graph's nodes with either positive or adverse edge weights on their edges. It operates by first transforming the graph by adding fresh node and edges without any weight, then applying the Bellman-Ford algorithm on transformed chart to determine the node’s size, and finally using Dijkstra's technique, you can figure out the shortest route between any two nodes. The time complex of the algorithm is O ($|V|$ 2log$|V|$ + $|V|$$|E|$), which makes it more efficient for large graphs \cite{aini}.

In a network with positive edge weights, the Dijkstra’s algorithm is a well-known method for figuring out the distance between any two source nodes and all other nodes. It operates by maintaining a queue of nodes and their shortest path estimates, choosing the node that has the lowest estimate and loosening its outward edges \cite{abous}, \cite{zhao}.

Path generation algorithms are computational methods utilised to connect two or more points in a network, and discover the shortest route, taking into account various constraints such as node and margin weights. In a graph with adverse edges, the Bellman-Ford technique is used to find the path between the origin node and every other node. The shortest path between any pair of vertices in a network with positive or adverse edge weights is found using the Floyd-Warshall method. A graph with either positive or adverse edge weights can be solved using the Johnson algorithm to find the shortest path between any two adjacent nodes, while in a network with positive edge weights, Dijkstra's technique is used to find the direct route between the source node and all other nodes.

\subsection{Overview of Nature-Inspired Optimization Algorithms}

A class of optimisation algorithms, swarm intelligence algorithms are modelled after how social organisms act in groups such as insects, bees, and birds. These algorithms utilise an abundance of simple agents, known as particles, ants, or wolves, that interact with one another and their surroundings to discover the ideal response to a specific issue. The application of swarm intelligence algorithms to a variety of optimisation problems, including path planning, scheduling, and machine learning, has been fruitful \cite{wahab}.

A prominent algorithm for swarm intelligence is ACO. The conduct of genuine ants is the basis for ACO, which leave pheromone traces to indicate the fastest way to get from their colonies to a source of food. ACO begins with a collection of artificial ants that investigate the search space and leave pheromone traces to indicate the quality of solutions. The insects then follow the pheromone traces, with a higher probability of following the pheromone-richer trail. As the insects continue to investigate, the pheromone traces are updated in accordance with the quality of the solutions discovered. The pheromone traces converge over time on the optimal solution \cite{wahab}.

PSO is based on the actions of avian colonies, which coordinate their movements in order to locate sustenance and avoid predators. Each particle in PSO represents a potential optimisation solution and traverses the search space based on its location and speed at the moment. The particles are affected by the most successful outcome that any particle has yet found, known as global best, as well as the optimal solution discovered by each element, known as individual best. As particles move through the search space, their positions and velocities are updated based on global and individual bests \cite{wahab}.

Another more modern swarm intelligence method is called Grey Wolf Optimizer (GWO). GWO is founded on grey wolves' coordinated hunting techniques, which are used to capture prey. Each wolf in GWO represents a potential optimisation solution and is categorised as beta, delta, or alpha based on performance value. Other wolves modify their positions and movements based on the beta, delta, and alpha wolves, with alpha wolf providing the most effective solution thus far. The search continues until either the optimal solution is located or a predetermined halting criterion is met \cite{wahab}.

In conclusion, swarm intelligence algorithms are a category of optimisation algorithms motivated by social animal behaviours in its entirety. These algorithms involve a large number of basic agents interacting with one another and their surroundings to discover the optimal solution to a given problem.

\subsection{Comparative Analysis of Algorithms}

Floyd-Warshall, Dijkstra's, Bellman-Ford, and Johnson algorithms have the following time, storage, and computation complexity characteristics,

Dijkstra's Method:

• With a binary heap implementation, the time complexity is O(($ |V| $ + $ |E| $)log$ |V| $), whereas O($ |V| $2) for simple array implementation.

• O($ |V| $ + $ |E| $) is storage complexity. 

• Complexity of computation: it requires a priority, which can be implemented using a binary heap or Fibonacci heap \cite{zhao}.

Bellman-Ford Methodology:

•	Time complication: O($ |V| $$ |E| $)

•	Storage difficulty: O(V)

•	Computational complexity: it requires a single array to store the minimum distances from the origin node to other nodes in graph \cite{zhao}.

Floyd-Warshall Method:

•	Duration complexity: O($ |V| $3)

•	Storage constraint: O($ |V|$2)

•	Computational complexity: storing the shortest distances between all pairs of nodes in the graph necessitates a two-dimensional array \cite{zhao}.

Johnson’s  Method:

•	O($ |V| $ 2log$ |V| $ + $ |V| $$ |E| $) is the time complexity.

•	O($ |V| $ + $ |E| $) is storage complication.

•	Computational complexity: It requires the Bellman-Ford algorithm to reweight the vertices and then Dijkstra's algorithm to determine the shortest path for each node \cite{aini}.

When combined with a binary heap or a Fibonacci heap, Dijkstra's has the most effective temporal complexity, whereas the Bellman-Ford is the one with the greatest time complexity. Floyd-Warshall and Johnson algorithms have comparable time complexity, but Floyd-Warshall algorithm requires more storage space than Johnson algorithm \cite{zhao}.

In terms of storage complexity, the Bellman-Ford algorithm requires the least storage space, whereas the Floyd-Warshall algorithm requires the most \cite{zhao}.

In terms of computational complexity, Dijkstra's and Johnson's algorithms require a priority queue to store the vertices, whereas the Bellman-Ford and Floyd-Warshall algorithms require only a single array and a two-dimensional array, respectively \cite{zhao}.

Based on only the complexity factor, the required algorithm cannot be chosen. Table I shows a comparison and visual representation of the benefits and drawbacks of Bellman-Ford, Floyd-Warshall, Dijkstra's, Johnson, and SI algorithms.

\begin{table}[!t]
\caption{Comparison of the Algorithms' Benefits and Drawbacks
\label{tab:table1}}
\centering
\resizebox{3.5in}{!}{
    \begin{tabular}{|c|>{\raggedright\arraybackslash}p{2in}|>{\raggedright\arraybackslash}p{2in}|}
        \hline
        \multicolumn{1}{|c|}{\textbf{Algorithm}} & 
        \multicolumn{1}{|c|} {\textbf{Advantages}} & \multicolumn{1}{|c|}{\textbf{Disadvantages}} \\
        \hline
        \centering Dijkstra & $\rightarrow$ Finds the fastest route in weighted graph. \newline $\rightarrow$ Find the optimal solution if all edge weights are non-negative. & $\rightarrow$ Cannot handle negative edge weights. \newline $\rightarrow$ Slow for large graphs.\\
        \hline
        \centering Bellman-Ford & $\rightarrow$ Can handle edge weights that are not positive. \newline $\rightarrow$ Can identify cycles that are negative in weight. & $\rightarrow$ Can have slow convergence if there are many weight cycles. \newline $\rightarrow$ Slow for large graphs. \\
        \hline
        \centering Floyd-Warshall & $\rightarrow$ Identifies the graph's shortest path between each pair of vertices. \newline $\rightarrow$ Accepts edge weights that are negative. & $\rightarrow$ Slow for large graphs. \newline $\rightarrow$ Requires a lot of memory for large graphs. \\
        \hline
        \centering Johnson & $\rightarrow$ Can handle negative edge weights. \newline $\rightarrow$ Can detect negative weight cycles. \newline $\rightarrow$ Faster than Bellman-Ford for sparse graphs. & $\rightarrow$ Requires pre-processing step to compute vertex weights. \newline $\rightarrow$ Slow for dense graphs. \\
        \hline
        \centering ACO & $\rightarrow$Can handle complex and dynamic search spaces. \newline $\rightarrow$ Robust to noise and uncertainty. \newline $\rightarrow$ Can find global optima. & $\rightarrow$ Sensitive to the initial conditions. \newline $\rightarrow$ A propensity to reach local optimum. \\
        \hline
        \centering PSO & $\rightarrow$ Simple and straightforward to apply. \newline $\rightarrow$ Can deal with complex optimisation issues. \newline $\rightarrow$ Can find global optima. & $\rightarrow$ Inclination to reach local optimum. \newline $\rightarrow$ Sensitive to parameter settings. \\
        \hline
        \centering GWO & $\rightarrow$ Fast convergence speed. \newline $\rightarrow$ Can handle non-linear and non-convex optimization problems. \newline $\rightarrow$ Can find global optima. & $\rightarrow$ Sensitive to initial conditions. \newline $\rightarrow$ Relies on the fitness function. \\
        \hline
    \end{tabular}
    }
\end{table}

\section{Literature Review }
The Preferred Reporting Items for Systematic Reviews and Meta-Analyses (PRISMA) methodology is done to select the important research and review articles based on the scope of the work using predefined criteria. It is for faster and in-depth analysis of articles as per the need. 
A total of 71 articles is reviewed as part of the research and considered 30 articles specifically for the review by using the PRISMA. Fig. 1 is a flowchart to depict the entire process of the review.

\begin{figure}[h]
\centering{\includegraphics[width=3.4in]{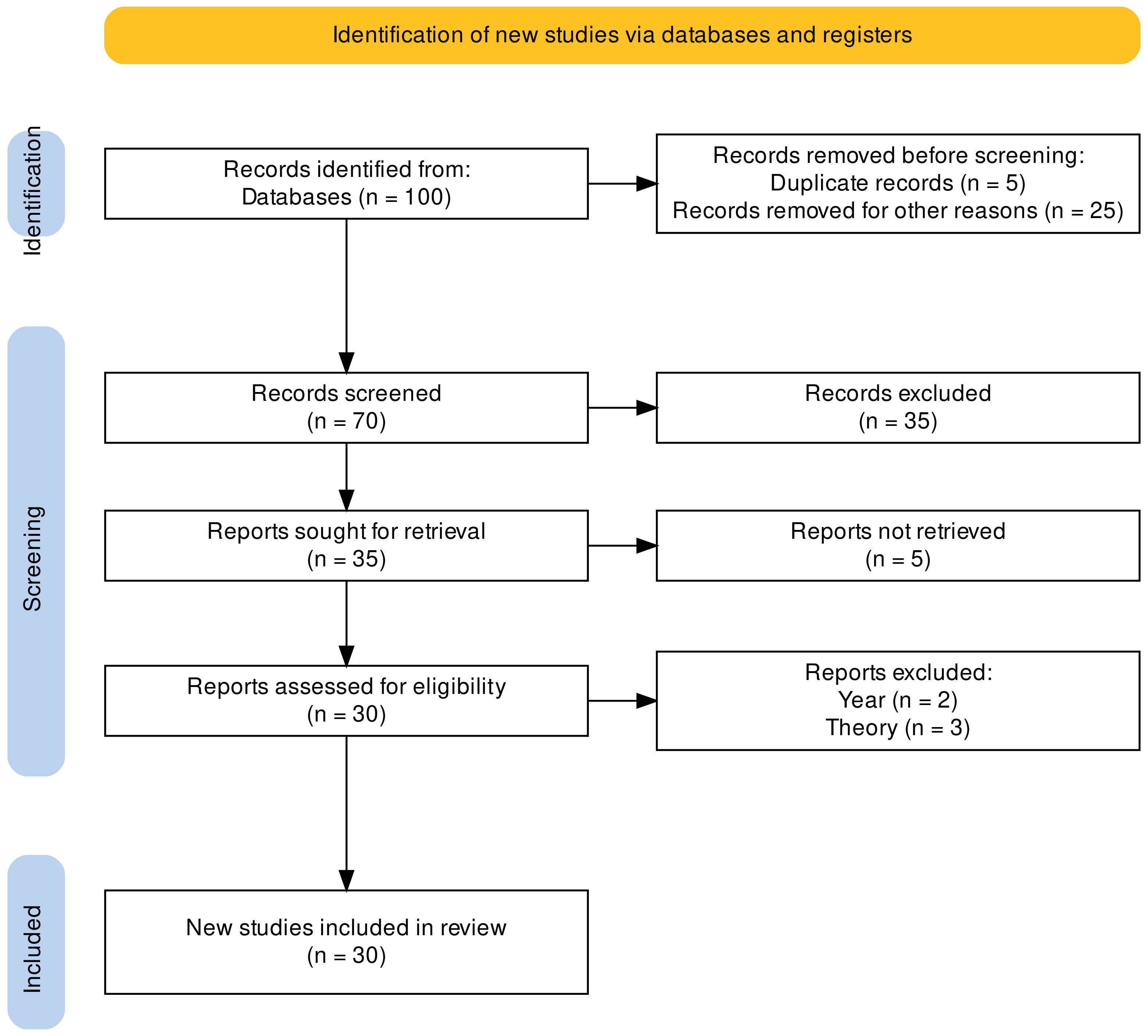}}
\captionsetup{justification=centering}
\caption{PRISMA Flowchart \cite{b42}}
\end{figure}

It is suggested that a modified ant algorithm be used as a quick and efficient way to plan UCAV paths. The planned route can ensure that the UCAV travels along the best route, with the least chance of getting lost and the least amount of energy expended, to its destination. The original ant algorithm can easily converge to local optimal solutions since the search space is large, but the search performance is slow. The speed of the selection process can be increased while the likelihood of local optimal solutions is decreased using the modified ant algorithm. The simulation results demonstrate that the proposed path-planning scheme can obtain an optimisation path that can be re-optimized when unanticipated hazards emerge \cite{mou}.

The creation of the origin-destination (OD) matrix, a key piece of information in vehicle routing or TSP, is discussed. To create an OD matrix, the shortest pathways between various nodes must be determined. Candidate techniques include the repetitive application of Floyd-Warshall method, Dijkstra's method, and other algorithms created specifically for this use. OD matrices can be computed on actual road networks using a variety of shortest path algorithms. Dijkstra's method with a rough container data structure fared well in the majority of the tested networks \cite{kim}.

A straightforward and widely used approach for finding the shortest paths between any pair of edges in an edge-weighted oriented graph is the Floyd-Warshall algorithm \cite{hougardy}. The single objective shortest path problem is extended to the multi-objective shortest path problem, which looks for effective connections between two pieces of a network while also resolving several competing objectives. MOSP is one of the most significant network optimisation problems, with widespread applications in the telecommunications, transportation, and project management industries. This work provides a multi-objective ACO-based approach for resolving the bi-objective short path issue. The results of the set of example problems indicate that the proposed algorithm yields non-dominated solutions of high quality and saves computation time \cite{ghoseiri}.

One of the most well-known techniques for finding the cheapest route between any pair of nodes in a network with at least one cycle is Floyd-Warshall. Additionally, we show that our algorithm's base is much easier to understand, which could be useful for instructional reasons. A brief example validates and demonstrates the implementation of our algorithm \cite{aini}. A comparison between the rectangular algorithm and the Floyd Warshall methodology is done. The Floyd Warshall algorithm performs marginally better for dense graphs, whereas the rectangular algorithm is more effective for sparse graphs \cite{singh}. Numerous swarm optimisation algorithms, from Adaptive Computation to the most recent, Grey Wolf Optimisation, have been developed since the early 1960s. In experiments, 30 well-known test functions are used to compare algorithms to one another and to provide a concise explanation of each algorithm. The results indicate that Differential Evolution (DE) and Particle Swarm Optimisation (PSO) are superior to all other approaches considered \cite{wahab}.

Most analysts predict that demand for electric and plug-in electric vehicles will increase. This need is fuelled by energy dependence, environmental concerns, and fluctuating fuel prices. These algorithms seek to reduce the distance or travel time between two locations. To overcome these optimisation difficulties, Dijkstra or Dijkstra-like algorithms are frequently employed. This problem necessitates alternative approaches. Energy-efficient navigation for EVs is a challenging challenge due to negative path costs produced by regenerative slowing, battery power and energy constraints, and vehicle parameters only known at query time. A deterministic model is Bellman-Ford optimisation technique it can solve routes containing negative trajectories. A representational model of electric vehicles is demonstrated. The model is then subjected to a Bellman-Ford search to find the route that uses the least amount of energy. The EV is then guided along the intended path using the developed solution. The efficacy of the Bellman-Ford algorithm is then investigated by employing the implemented algorithm to maps of varying sizes \cite{abous}.

Given is a description and implementation of the author's heuristic approach for finding the optimal path in between two selected graph edges for a heuristically estimated function \cite{akmay}. In order to show customers, the quickest route to take in order to buy the items on their shopping list, the study applies three shortest-path-finding algorithms to a grocery store: Floyd-Warshall, Bellman-Ford, and Dijkstra's Algorithms. The created application is a server-client system with a C programming running on MS VS Environment as the server and a tablet app for Android as the client. In most circumstances, the three techniques give the same path lengths, but depending on the number of items in the list, they take different amounts of time to compute. Actual experiments were conducted in a grocery store, and after using the application, the user's purchasing time, distance travelled, and number of steps were all reduced \cite{delacruz}.

A novel and effective approach to the Dijkstra, Bellman Ford, Floyd-Warshall, and Viterbi algorithms is offered through parallel programming, analysis of the results from various tests, and comparison of these searching tactics on graph systems. This study has the potential to generate new approaches to these strategies and new discussions regarding the need to enhance the old algorithms \cite{popa}. The Exploration Control Particle Swarm Optimisation technique is suggested as a solution to the grid job scheduling problem; it is suggested to create a velocity update rule depending on exploration capability; and the retrieval behaviour of particles is controlled by the exploration control probability (ECP) parameter, which increases search effectiveness. In addition, a unique local search is proposed as a means of enhancing the search's efficacy. The proposed method has the benefits of preserving exploration during the search process, being easy to implement, lowering complexity and computing time. Scheduling problems for grid tasks can be successfully handled using ECPSO \cite{chen}.

Utilising an ambulance outfitted with the necessary medical personnel and supplies is one of the first ways to assist the victim or patient. First aid may be administered to victims and patients more easily if an ambulance is available and there is reliable information about the victims and the state of the roads. Determining the finest (closest and shortest) path to the closest hospital will enable patients to receive supportive care. The Dijkstra algorithm is used in this application to determine the fastest path to the closest hospital. The Floyd-Warshall algorithm is used to determine the nearest distance to the hospital. This application is intended to aid in the initial management of the victim or emergency patient by supplying the ambulance calling report and determining the optimal route to the closest hospital \cite{risald}. The single-source shortest path is a popular approach in graph analysis. A straightforward approach called Bellman-Ford can be used to find the shortest route to a source node, allowing the identification of adverse-weighted cycles in a graph. Additionally, it represents a class of sequential algorithms that distribute tasks unevenly and with data dependencies. Many algorithms have recently been implemented using graphics processors, which have also been utilised as supercomputer accelerators. In this study, we use CUDA to accelerate the Bellman-Ford algorithm on the GPU, making it faster than earlier algorithms in traversing compact and shallow graphs \cite{nazarifard}.

This article describes in detail the algorithmic procedure of three prevalent fastest path algorithms, namely Floyd-Warshall, and Floyd-Warshall. It describes the execution stages of the three algorithms using case diagrams. This paper contrasts three algorithms based on their application range, geographical difficulty, time complexity, and negative weight and reaches a conclusion. When selecting the shortest path algorithm, it is necessary to take actual conditions and the characteristics of each algorithm into consideration \cite{zhao}. We look at a digraph with allowed negative edge costs and the single-source shortest path. To improve the Bellman-Ford algorithm's running time bound for networks with an uneven distribution of negative cost edges, a hybrid of the two algorithms is recommended. If there is a shortest path to every vertex that can be reached from the source with no more than k adverse cost edges, the problem can be solved in no more than k 2 iterations. In addition, a new, uncomplicated demonstration suggests that the Bellman–Ford algorithm generates a shortest paths tree \cite{dinitz}. Graph travelling problems are among the earliest network theory problems. Numerous dynamic programming problems with discrete state and discrete time are among the many applications for the shortest path techniques, as are network optimisation issues involving road and telecommunications networks. Currently, graphs offer a straightforward and frequently helpful formal model of biological networks that reflects one-to-one interactions between biological elements. The several kinds of shortest path finding algorithms are briefly described. To demonstrate how each algorithm works, examples of the C implementations of the algorithms under consideration are provided. The evaluation outcomes of the algorithms and their time complexity are displayed \cite{sapundzhi}.

Prior to travelling, one of the most essential considerations is determining the travel route, particularly the shortest route. The quickest route was found using both the traditional and heuristic methods. These two approaches will be compared to determine which yields the greatest outcome, the Floyd-Warshall method considers every possible path so that some routes are revealed, reducing the amount of time needed for searching. Floyd-Warshall method provides a superior solution based on the testing conducted, with a mileage of 22.7 km compared to 24.8 km for the greedy algorithm. This result indicated that an extended time is necessary because it takes into account the distance to all points \cite{azis}. As was the case in the Bandung area, notably in the sphere of tourism, where congestion hampered visitor activities, congestion in a region can result in ineffective utilisation of assets and can have a substantial impact on the flow of socioeconomic activity in the region. It requires an application that can determine the optimal route to assist travellers in reaching their tourist destination and avoiding highway congestion \cite{laga}. Using numerically derived location information, path finding algorithms enable the discovery of routes that could be taken between the required source location and the intended destination point. A significant issue in the field of study is finding the fastest and most accurate shortest path between two points, especially in situations that are constantly changing. Path tracing algorithms are used to tackle a wide range of issues in a variety of fields. Each path-tracing algorithm was evaluated utilising a distance-based graph of the provinces of Turkey. Java programming language was utilised to evaluate each algorithm. Java's open-source JFX map library was utilised to depict the route between the starting and ending points. It is anticipated that large, graph data would be used in next studies. Additionally, graph data for province centres and district centres will be compiled and tested with each algorithm \cite{alkan}.

In order to reduce pickers' journey time in manual warehouses, a article suggests a brand-new metaheuristic routing method. This algorithm is known as FW-ACO and is based on the ACO metaheuristic, which is merged and blended with the Floyd-Warshall (FW) algorithm. First, the effectiveness of the algorithm's solutions to the selection problem are assessed in relation to the default values for the key ACO parameters. Second, the effectiveness of the FW-ACO method is evaluated against six other algorithms that are frequently applied to optimise pick-ers' journey distances. The results show that the FW-ACO is a promising algorithm that frequently finds accurate solutions and is generally able to give results that are superior to those of heuristic and metaheuristic algorithms. The FW-ACO algorithm also exhibits a very efficient computation time, which qualifies it for real-time picker route specification. The FW–ACO algorithm is ultimately implemented in a real-world case study with constraints on the order of item selection in order to demonstrate its practical utility and quantify the resulting cost reductions \cite{santis}.

ACO, a type of traditional swarm intelligence algorithm, is particularly good at solving combinatorial optimisation issues. The research proposes a novel enhanced pheromone update strategy to further increase convergence speed while maintaining solution quality. Using changing data to optimise the path, this system reinforces the pheromone at the edges. When the ACO algorithm nears a pre-set stagnation stage, a novel pheromone- smoothing technique is also developed to reinitialize the pheromone matrix. The improved algorithm is examined and examined against a set of sample test cases. The experimental results demonstrate that the enhanced ant colony optimisation algorithm outperforms competing algorithms in terms of both the diversity of the obtained solutions and convergence speed \cite{ning}. Methodologically, the direct comparison method was applied. To support a map retrieved from Google Maps, computations were made using graph concept and the Haversine Formula. Both methods discovered optimal path. The outcome indicated that Dijkstra's algorithm can generate shorter paths than the Bellman-Ford Algorithm \cite{pramudita}.

By classifying satellite images from ISPRS using a Deep Convolutional Encoder-Decoder architecture-Seg Net, researchers have improved the accuracy and effectiveness of optimal path planning. The cost map is utilised to find an energy-efficient path using modified gain-based ant colony optimisation. MGACO outperforms existing algorithms in terms of runtime and path length, according to a comparison with contemporary algorithms \cite{sangeetha}. A comparison of the Floyd-Warshall and Dijkstra algorithms is used to find the best railway route. The optimal route is the one with the least expensive train fare. The results of route discovery will be shown in a PHP-based web application that is supported by a MySQL database. Four parameters are used to compare the outcomes of these two algorithms: Level of completion, time complexity, memory complexity, and level of optimisation. Based on our experiments, the Dijkstra algorithm performs better than the Floyd-Warshall algorithm on these four parameters \cite{dermawan}. In this article, we discuss the most popular graph algorithms utilised in a range of fields, including computational biology and transportation networks. We will first look at the DFS and BFS traversal algorithms, which are both well-known. Then, we look at the issue of finding the shortest path between two edges and discuss the Dijkstra, Bellman-Ford, and Floyd-Warshall. We will also look at another foundational graph issue, the Ford-Fulkerson algorithm for computing the maximal flow in a graph. After that, we will examine two well-known algorithms—the Kruskal method and the Prim algorithm—for computing the minimal spanning tree of a given graph. Finally, we will examine the travelling salesman problem and the nearest neighbour algorithm, which is a heuristic for this issue \cite{dondi}.

The high mobility of UAVs has not been able to solve the problems of target localization and identification, despite the fact that there are many research ideas on the path planning concerns of UAVs in the literature. For various mission-critical UAV operations, optimal decisions must be made in order to address these issues in UAV path planning. The goal of path planning approaches is to avoid colliding with other UAVs while still determining the best and quickest route. It is essential to have techniques for calculating the safest route to an ultimate destination in the shortest amount of time. The diverse path planning strategies used by UAVs are categorised into three primary groups in this paper: representational strategies, cooperative strategies, and non-cooperative strategies. The existing options have undergone a thorough review based on each area of UAV path planning \cite{aggarwal}.

Farmers incur substantial financial losses due to orchard damage caused by birds. So, a Particle Swarm Optimisation (PSO)-based optimisation algorithm for UAV path planning is developed. This approach was used since an entrance optimisation technique was required to start the initial tests. PSO is an uncomplicated algorithm with few control parameters that maintains high performance. This path planning optimisation algorithm employs optimisation and stochastic techniques to control the drone's flying time and distance in order to overcome the limitations of existing systems. The performance of the proposed algorithm was evaluated using a simulation of all possible tree cases \cite{mesquita}.

Travellers face challenges due to traffic, distance, and the abundance of tourist locations. The Simple Additive Weighting (SAW) technique is used to examine the data on congestion and distance. The algorithm assigns the smallest weight to the selection of the optimal route, which is 5.127. Testing has revealed that the typical estimate time for two to five tourist destinations is three to five seconds. It is anticipated that this application will provide optimal solutions for the selection of tourist travel routes \cite{arfananda}. The UAV mission planning system must decide the best way through a difficult terrain, and one of its key components is the UAV path planning problem. HSGWO-MSOS, a novel hybrid algorithm is proposed as a solution to this problem. To speed up convergence while maintaining the population's capacity for exploration, the GWO algorithm's phase is condensed. To improve the ability to exploit, the commendation phase of the SOS algorithm gets altered and combined with the GWO. The simulation experimental results demonstrate that the HSGWO-MSOS algorithm can effectively acquire a feasible and efficient route, and that its performance is superior to that of the GWO, SOS, and SA algorithms \cite{qu}.

RLGWO, a novel algorithm for solving this problem. Reinforcement learning is incorporated into the proposed algorithm so that the person is managed to transfer tasks in accordance with their cumulative performance. The simulation results demonstrate that the RLGWO algorithm can effectively acquire a feasible and effective route in complex environments \cite{gai}. In order to efficiently distribute products to consumers, a business must optimise the shortest route. This optimisation can assist businesses in optimising their mileage and expenses from origin to destination. For each vertex it passes through, the Floyd-Warshall algorithm can compare all feasible graph paths with the least number of iterations. Based on the outcomes of this study's calculations, the shortest route between Mataram and Praya was determined to be 31 kilometres \cite{febryan}.

A network model with a complex topology is implemented using the modified Floyd-Warshall approach, which removes constraints brought on by negative weights and cycles and automates the computation of shortest path \cite{sakharov}. Greater accuracy and efficiency in finding the shortest paths are required due to the growing volume of data provided by more complex network connections. As a result of their recognised high calculation time complexity, the traditional precise methods of the past are no longer ideal for large-scale data processing. The distributed mode of the enhanced SDM algorithm greatly speeds up execution compared to Dijkstra's and Floyd's algorithms. The data test demonstrates that the new algorithm enhances the processing of massive amounts of data \cite{yuan}.

For the purpose of resolving the multi-objective path-planning problem in a radioactive environment, an enhanced ant colony optimisation technique is put forth. The IACO algorithm has a better likelihood than the PSO and CPSO algorithms of locating the optimal route, according to five simulations of experimental data. In addition, it was more effective than ACO at determining the path. IACO can help personnel reduce their exposure to radiation, and experimental simulations have proven its efficacy and validity in solving the multi-objective inspection path-planning problem in a radioactive environment \cite{xie}.

The expanding use of USVs in several applications in challenging situations has called for the creation of unique path planning techniques that can successfully handle path planning issues with multiple objectives. In order to broaden the application of ACO to meet a variety of goals, this work focuses on applying fuzzy inference systems. In terms of solution quality, the results revealed that ACO with Mamdani beat other approaches, whereas ACO with RMSE exhibited a faster rate of convergence and ACO with TSK achieved a better balance between convergence speed and solution quality \cite{ntakolia}.

The Toba Regency Government prioritises the development of the tourism industry because the Toba Regency's natural resources have tremendous potential. The purpose of this study is to calculate the duration and quickest route to a well-known tourist attraction. The fastest route is decided using the Dijkstra and Floyd - Warshall and Dijkstra’s methods. Floyd-Warshall evaluates all potential graph paths for each vertex, whereas Dijkstra's approach compares each solution set to find the best one. Based on the research findings, both algorithms generate the same shortest route for 24 tourist attractions in Toba, and a weighted graph is produced \cite{samosir}.

In networking, logistics transportation, and other fields, the problem of travelling salesmen is frequently used as a typical illustration of an NP-complete issue. This problem is solved using Discrete Lion Swarm Optimisation. First, DLSO introduces discrete coding and order crossover operators. Second, to enhance the power of local search, C2-opt method is employed. Then, a Parallel Discrete Lion Swarm Optimisation is suggested to boost the algorithm's performance. Diverse populations exist within the PDLSO, and each subpopulation independently executes DLSO in parallel. Experiments on some TSP benchmark problems demonstrate that the DLSO algorithm is more precise than other algorithms, and that the PDLSO algorithm can effectively reduce the running time \cite{daoqing}.

Planning the ideal route for urban vegetation is a technique to maximise its utilisation, which is advantageous for both the visitors' path-finding and the physical and emotional well-being of city dwellers. Street view images of Osaka City is obtained using Google Street View. The street view photos are segmented, and GVI) of the city is evaluated using the semantic segmentation model. GVI's ideal path based on the GVI is determined using the Floyd-Warshall method and adjacency matrix, circumventing the ArcGIS software's drawbacks. By combining all the information, we can perform an intuitive and unbiased analysis of the street-level vegetation in the study region \cite{hu}.

The Travelling Salesman Problem (TSP) is solved via a discrete cuckoo search based on cluster analysis and random walk. The former is used to maintain the superior answers that the algorithm has found, whilst the latter is used to maintain the population's diversity. The k-means approach divides cities into k optimisation categories for large-scale TSP issues, and random procedures are utilised to combine them. The local optimisation operator, which boosts the algorithm's convergence rate, is a simple 2-opt operator. Experimental results show that the suggested algorithm's stability and precision are superior to those of the most recent optimisation techniques \cite{yang}. In complicated situations, the QLTSO algorithm, a tuna swarm optimisation method based on reinforcement learning, can plan safe and reliable AUV navigation pathways with enhanced convergence and robustness. The planning success rate of the algorithm can reach 100 \%, and simulation results show that it can provide safe and reliable AUV navigation paths with increased convergence and robustness \cite{wu}.

\section{Methodology}

The objective of the study is proposed in accomplishing the following targets:

\begin{itemize}
    \item Comparative analysis of Bellman-Ford, Floyd-Warshall, Johnson algorithm and ACO, PSO, GWO algorithm, to choose the effective algorithm for the application.
    \item Implement the chosen path generation algorithm, compare its results with that of general existing algorithm and show the effectiveness.
    \item Give an insight about the future developments of the proposed model.
\end{itemize}

First, research and review papers are collected from Elsevier, IEEE, Springer, MDPI and other databases. A brief study on 70 papers is done and further 30 papers are chosen based on pre-defined criteria like published year, theory, and model used using PRISMA method. 

Second, a detailed study was performed on 30 papers and made a comparitive analysis of ACO, PSO, GWO, Bellman-Ford, Johnson, and Floyd-Warshall algorithms. Following the analysis, Floyd-Warshall and ACO algorithm is chosen as the effective algorithm for the above-mentioned application. 

Then, the modified Floyd-Warshall with ACO method is been implemented using Jupyter Notebook (The Python Environment) and results are observed for different stages of implementation. The complete planning of the research is represented in the Fig. 2.

\begin{figure}[h]
\centering{\includegraphics[width=3in, height=7.5in]{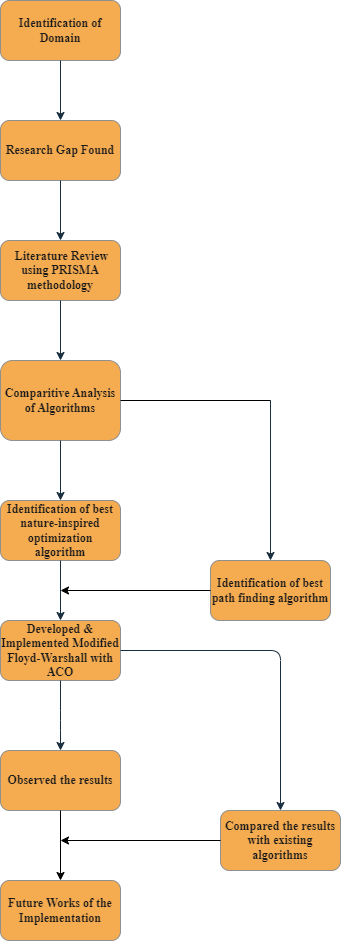}}
\captionsetup{justification=centering}
\caption{Research Framework}
\end{figure}

\section{Result and Discussion}

Based on the review and complexity factors, pros, cons of each algorithm. Floyd -Warshall algorithm \& Ant Colony Optimization (ACO) algorithm is considered over other algorithms for planning of above-mentioned transportation modes. The actual benefit of Integrating Floyd-Warshall algorithm with Ant Colony Optimization (ACO) helps in considering physical, social, economic, technical, and time constraints involved as part of various kinds of transportation planning. The model’s effectiveness would be improved when compared to traditional ACO, by optimizing the paths generated under the afore-mentioned constraints.

It implies that integration of these two algorithms leads to proper planning of route between two destinations. By implementing the modified Floyd-Warshall with ACO algorithm on quasi structured points, the first stage of implementation is completed. From the results observed, when compared to general Floyd-Warshall algorithm, the proposed method is effective and less complex. Implementation is divided into different parts and the results of all these parts are presented and explained below:

\begin{figure}[h]
\centering{\includegraphics[width=3.5in]{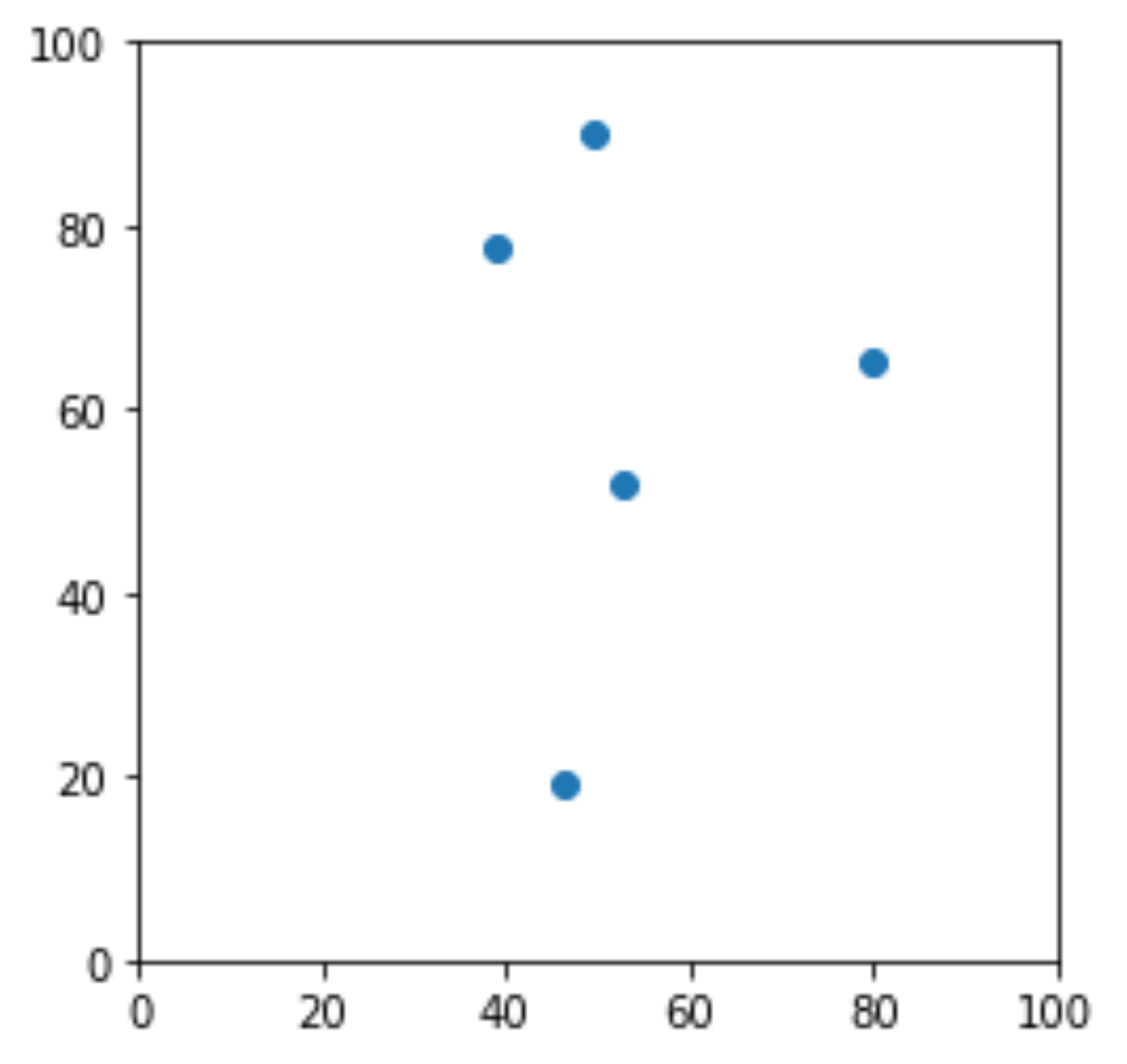}}
\captionsetup{justification=centering}
\caption{Organization of Quasi-Sturctured Points within a Boundary}
\end{figure}

As shown in Fig. 3, the points are generated and placed at random locations within the prescribed boundary region making it appear like quasi-structured. The number of points or the boundary limits can be changed according to our necessity. In Fig. 4, all path connections are made between the points using the path generation algorithm of point connections.

\begin{figure}[h]
\centering{\includegraphics[width=3.5in]{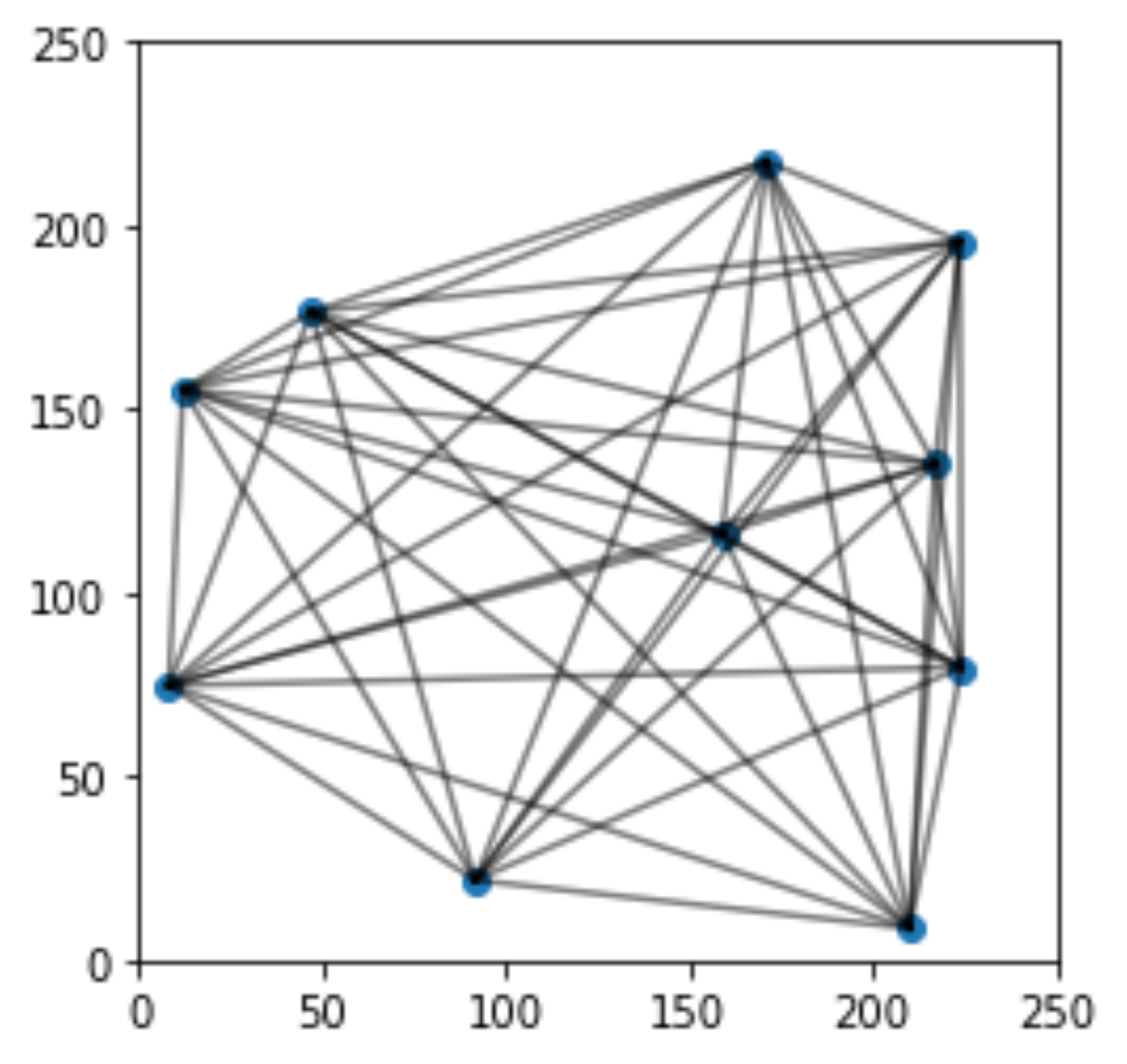}}
\captionsetup{justification=centering}
\caption{Paths between points generated using the path generation algorithm}
\end{figure}
\begin{figure}[h]
\centering{\includegraphics[width=3.5in]{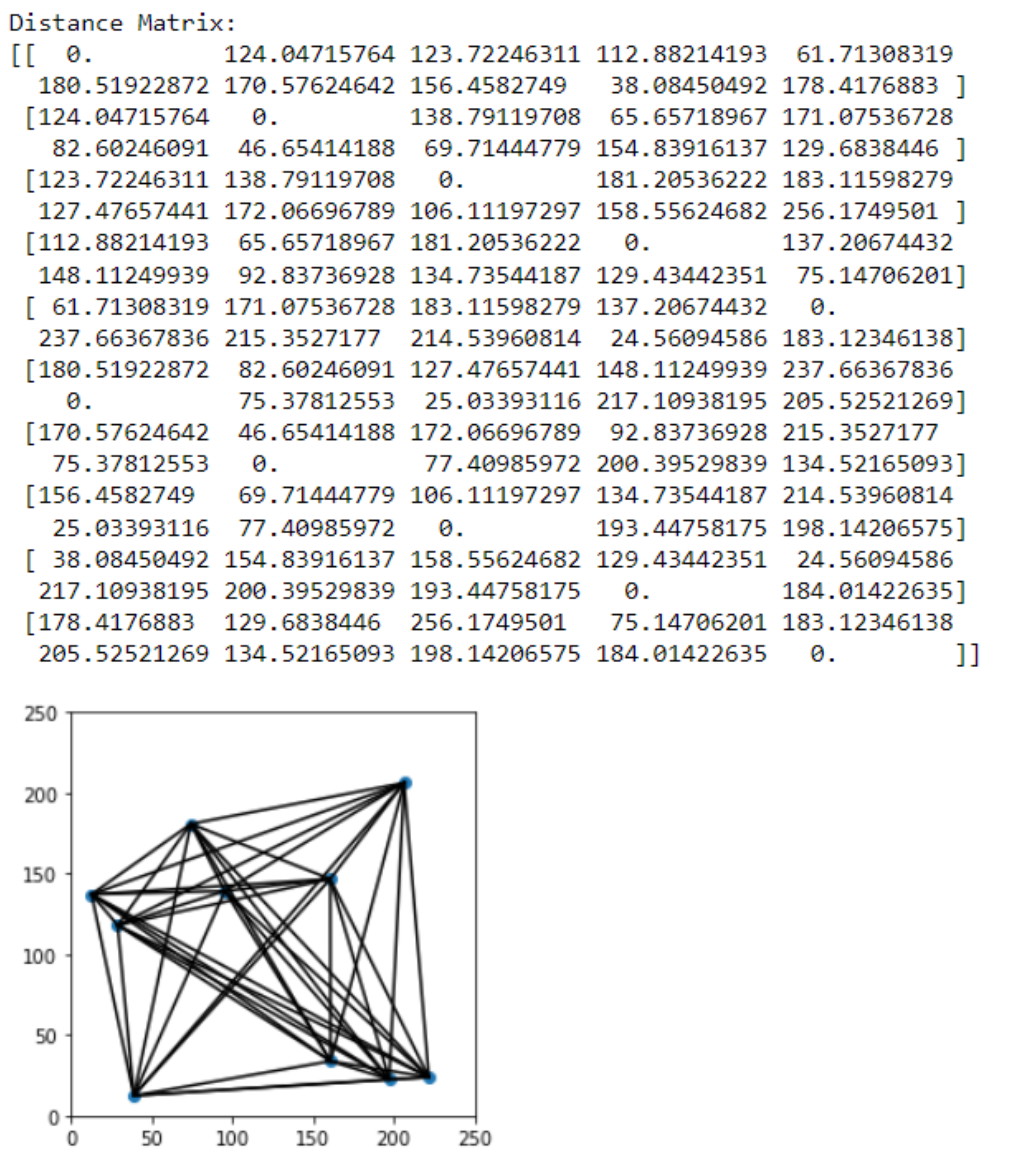}}
\captionsetup{justification=centering}
\caption{Paths between points and Distance matrix are generated using Floyd-Warshall algorithm}
\end{figure}

But in Fig. 5, all the points are connected but only all the possible shortest paths are generated using Floyd-Warshall method. Also, distance matrices are used to express the list of all shortest paths. The connections formed in Fig. 5 are showed in a simplified form in Fig. 6, by using 5 points in same boundary region and the points through which path traversed is shown using a graph legend.

\begin{figure}[h]
\centering{\includegraphics[width=3.5in]{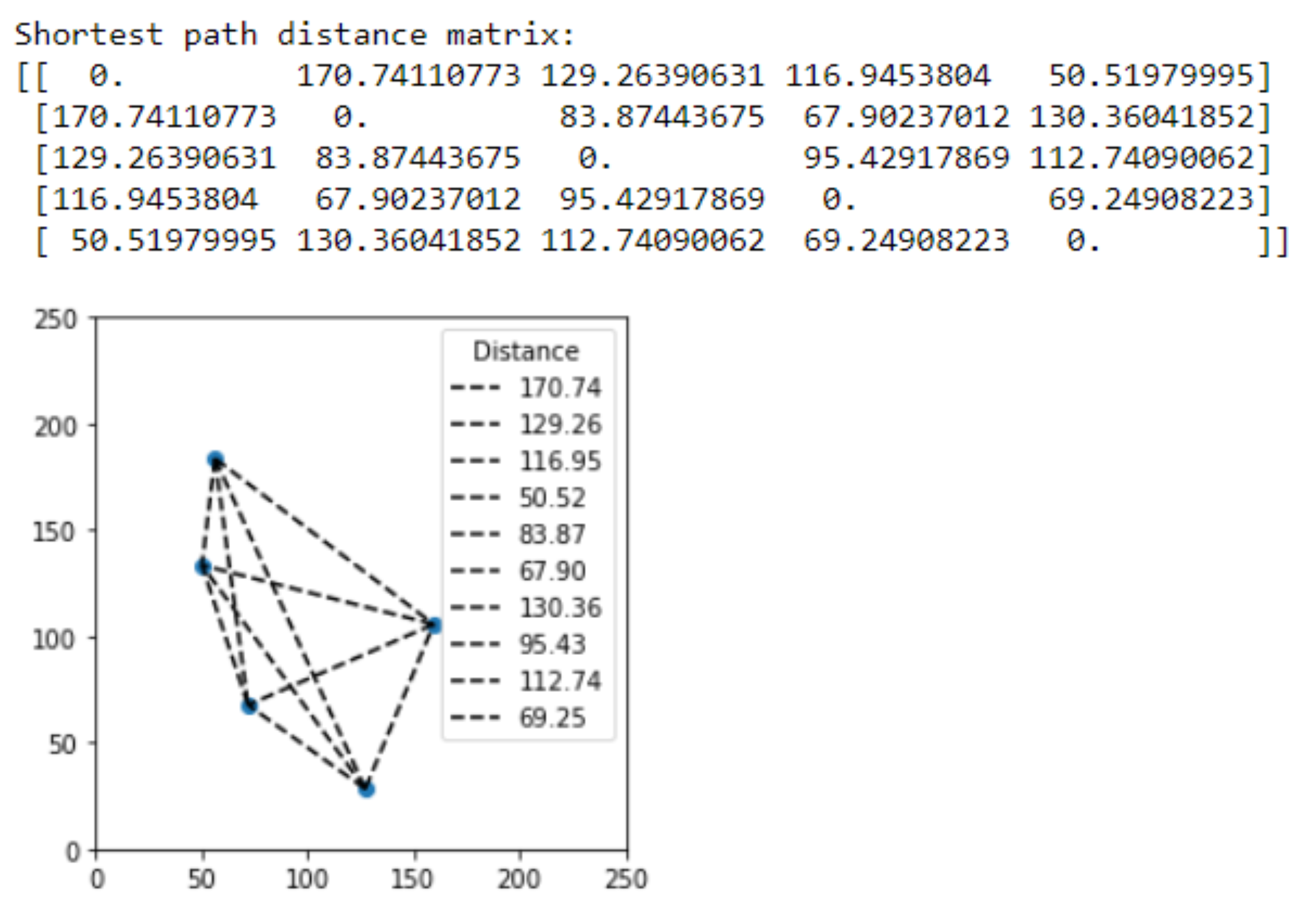}}
\captionsetup{justification=centering}
\caption{Step-by-Step approach of Floyd-Warshall Algorithm}
\end{figure}
\begin{figure}[h]
\centering{\includegraphics[width=3.5in, height=4.5in]{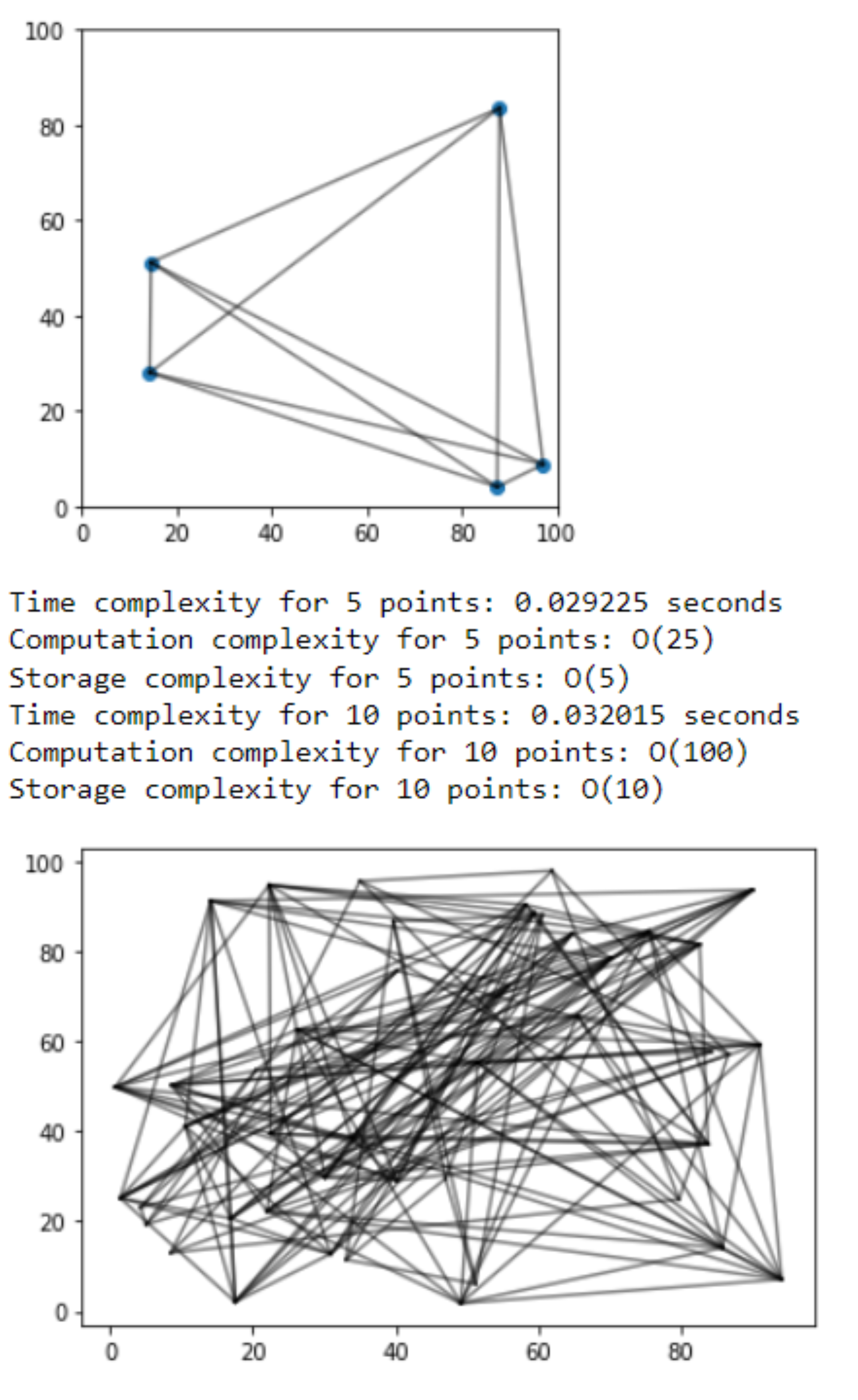}}
\captionsetup{justification=centering}
\caption{Complexity Metrics of Floyd-Warshall Algorithm}
\end{figure}

The general Floyd-Warshall’s complexity for time, storage, and computation is shown in Fig. 7, with respect to 5 and 10 points in the same boundary region. At the same time, the complexity factors of the results obtained using proposed method is shown in Fig. 8. 

Out of the three complexity factors, time complexity shows a vast difference (refer TABLE II) using proposed algorithm. 

\begin{figure}[h]
\centering{\includegraphics[width=3.5in]{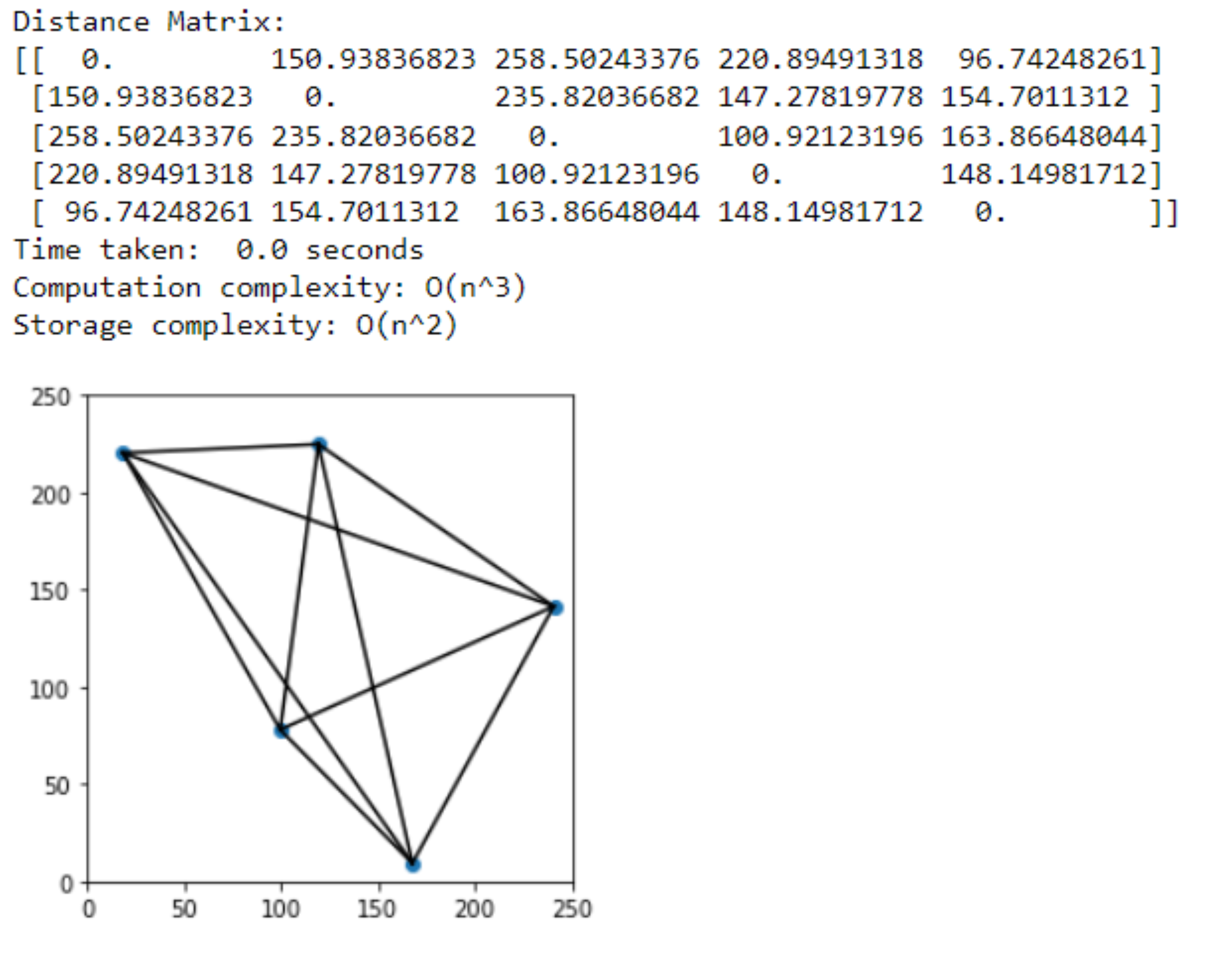}}
\captionsetup{justification=centering}
\caption{Complexity Metrics of Modified Floyd-Warshall with ACO Algorithm}
\end{figure}

\begin{table}[!t]
\caption{Time Complexity of Existing algorithm and Proposed algorithm
\label{tab:table1}}
\centering
\resizebox{3.5in}{!}{
    \begin{tabular}{|c|>{\raggedright\arraybackslash}p{2in}|>{\raggedright\arraybackslash}p{2in}|}
        \hline
        \multicolumn{1}{|c|}{\textbf{S. No}} & 
        \multicolumn{1}{|c|} {\textbf{Algorithm}} & \multicolumn{1}{|c|}{\textbf{Time Complexity (seconds)}} \\
        \hline
        \centering 1. & General Floyd-Warshall algorithm & 0.02 \\
        \hline
        \centering 2. & General Floyd-Warshall with ACO algorithm & 0.01 \\
        \hline
        \centering 3. & Modified Floyd-Warshall with ACO algorithm & \textbf{0.001} \\
        \hline
    \end{tabular}
    }
\end{table}

\section{Conclusion}

The initial model for planning various modes of transportation is developed using the Floyd-Warshall method and ACO. The first stage of implementation is completed using the proposed method and results are observed graphically. The model will be improved in the following two stages in the future:

\begin{enumerate}
    \item In this respective stage, basic structural constraints related to graphical paths are to be considered as conditions for ACO. With the help of this implementation, we can choose the optimal solution from a collection of shortest paths according to the prescribed requirement.
    \item To implement the developed model from above stage for planning of different modes of transportation. For this with help of GIS and geographical \& environmental constraints mentioned in the previous sections of this paper are considered to identify the best route between a network’s source and destination for any mode of transportation.
\end{enumerate}

The actual gain from creating the aforementioned model is to reduce the costs, expertise \& resource required, and time utilized for the above-mentioned planning by a vast difference.

\begin{IEEEbiographynophoto}{Hariram Sampath Kumar}
was born in Salem, Tamil Nadu, India in 2001. He received the B.E. degree in mechatronics engineering from Sona College of Technology (affiliated to Anna University, Chennai), Salem, Tamil Nadu in 2022. He is currently  pursuing the MTech degree in artificial intelligence at Amity University Noida, Uttar Pradesh, India.

His research interest includes development of field robots \& aerial robots, swarm intelligence, deep learning, and artificial intelligence for sustainable development.

Mr. Sampath Kumar has received Tamil Nadu State Council for Science \& Technology Grant under Student Projects Scheme 2021 – 2022.

\end{IEEEbiographynophoto}
\begin{IEEEbiographynophoto}{Archana Singh}
received the B.Sc degree in computer science from University of Delhi in 1997, Masters' degree in computer science and management from University of Pune, in 1999, MTech degree in computer engineering from IETE, New Delhi in 2008 and Ph.D. degree in data science focusing on data mining and machine learning from Amity University Noida in 2015.

She has 20 years of experience in academia and currently Professor \& Head - Department of Artificial Intelligence at Amity University Noida. Her research interests include artificial intelligence, machine learning, and image processing.
\end{IEEEbiographynophoto}
\begin{IEEEbiographynophoto}{Manish Kumar Ojha}
a faculty member at Amity University Noida, has extensive academic and professional experience in technology-driven engineering \& management. He received the BTech in Mechanical Engineering and MTech in Industrial Engineering and Management from IIT Kharagpur, India. He is currently pursuing the Ph.D. degree in Operations and Supply Chain Management from IIM Jammu, India. 

His research interests include Operations Research, Quantitative Methods, Supply Chain Management, and Optimization Techniques. He has also contributed significantly to academia, including publications in international journals and presentations at prestigious conferences. 

Mr. Ojha has also achieved notable achievements, including a DAAD Scholarship for a master's project at Technical University Dresden, Germany, and qualifying for CAT and GATE examinations.
\end{IEEEbiographynophoto}


\begin{thebibliography}{42}
\bibliographystyle{IEEEtran}

\bibitem{mou} C. Mou, W. Qing-xian, and J. Chang-sheng, “A modified ant optimization algorithm for path planning of UCAV,” \textit{Applied Soft Computing}, vol. 8, no. 4, pp. 1712–1718, 2008. doi:10.1016/j.asoc.2007.10.011 

\bibitem{kim} B.I. Kim and S. Jeong, “A comparison of algorithms for origin–destination matrix generation on real road networks and an approximation approach,” \textit{Computers \& Industrial Engineering}, vol. 56, no. 1, pp. 70–76, 2009. doi:10.1016/j.cie.2008.03.016

\bibitem{hougardy} S. Hougardy, “The Floyd–Warshall algorithm on graphs with negative cycles,”\textit{ Information Processing Letters}, vol. 110, no. 8–9, pp. 279–281, 2010. doi:10.1016/j.ipl.2010.02.001 

\bibitem{ghoseiri} K. Ghoseiri and B. Nadjari, “An ant colony optimization algorithm for the bi-objective shortest path problem,” \textit{Applied Soft Computing}, vol. 10, no. 4, pp. 1237–1246, 2010. doi:10.1016/j.asoc.2009.09.014 

\bibitem{aini} A. Aini and A. Salehipour, “Speeding up the Floyd–Warshall algorithm for the cycled shortest path problem,” \textit{Applied Mathematics Letters}, vol. 25, no. 1, pp. 1–5, 2012. doi:10.1016/j.aml.2011.06.008 

\bibitem{singh} A. Singh and P. Kumar Mishra, “Performance analysis of Floyd Warshall algorithm vs rectangular algorithm,” \textit{International Journal of Computer Applications}, vol. 107, no. 16, pp. 23–27, 2014. doi:10.5120/18837-0372

\bibitem{wahab} M. N. Ab Wahab, S. Nefti-Meziani, and A. Atyabi, “A comprehensive review of swarm optimization algorithms,” \textit{PLOS ONE}, vol. 10, no. 5, 2015. doi:10.1371/journal.pone.0122827

\bibitem{abous} R. Abousleiman and O. Rawashdeh, “A bellman-ford approach to energy efficient routing of electric vehicles,” \textit{2015 IEEE Transportation Electrification Conference and Expo (ITEC)}, Dearborn, MI, USA, 2015, pp. 1-4. doi:10.1109/itec.2015.7165772

\bibitem{akmay} D. A. Akmaykin \textit{et al.}, “Optimization of the search algorithm for the shortest route,” \textit{2015 International Conference “Stability and Control Processes” in Memory of V.I. Zubov (SCP)}, St. Petersburg, Russia, 2015, pp. 545-548. doi:10.1109/scp.2015.7342207 

\bibitem{delacruz} J. C. Dela Cruz\textit{ et al.}, “Items-mapping and route optimization in a grocery store using Dijkstra’s, bellman-ford and Floyd-Warshall Algorithms,” \textit{2016 IEEE Region 10 Conference (TENCON)}, Singapore, 2016, pp. 243-246. doi:10.1109/tencon.2016.7847998

\bibitem{popa} B. Popa and D. Popescu, “Analysis of algorithms for shortest path problem in parallel,” \textit{2016 17th International Carpathian Control Conference (ICCC)}, High Tatras, Slovakia, 2016, pp. 613-617. doi:10.1109/carpathiancc.2016.7501169 

\bibitem{chen} R.-M. Chen, Y.-M. Shen, and C.-T. Wang, “Ant Colony Optimization Inspired Swarm Optimization for Grid Task Scheduling,” \textit{2016 International Symposium on Computer, Consumer and Control (IS3C)}, Xi'an, China, 2016, pp. 461-464. doi:10.1109/is3c.2016.122

\bibitem{risald} Risald, A. E. Mirino, and Suyoto, “Best routes selection using Dijkstra and Floyd-Warshall algorithm,” \textit{2017 11th International Conference on Information \& Communication Technology and System (ICTS)}, Surabaya, Indonesia, 2017, pp. 155-158. doi:10.1109/icts.2017.8265662 

\bibitem{nazarifard} M. Nazarifard and D. Bahrepour, “Efficient implementation of the Bellman-Ford algorithm on GPU,” \textit{2017 IEEE 4th International Conference on Knowledge-Based Engineering and Innovation (KBEI)}, Tehran, Iran, 2017, pp. 773-778. doi:10.1109/kbei.2017.8324901 

\bibitem{zhao} L. Zhao and J. Zhao, “Comparison Study of Three Shortest Path Algorithm,” \textit{2017 International Conference on Computer Technology, Electronics and Communication (ICCTEC)}, Dalian, China, 2017, pp.748-751. doi:10.1109/icctec.2017.00165

\bibitem{dinitz} Y. Dinitz and R. Itzhak, “Hybrid bellman–ford–dijkstra algorithm,” \textit{Journal of Discrete Algorithms}, vol. 42, pp. 35–44, 2017. doi:10.1016/j.jda.2017.01.001 

\bibitem{sapundzhi} F. I. Sapundzhi, and M. S. Popstoilov, "Optimization algorithms for finding the shortest paths," \textit{Bulgarian Chemical Communications}, vol. 50, no. Special Issue B, pp. 115-120, 2018.

\bibitem{azis} H. Azis, R. Mallongi, D. Lantara, and Y. Salim, “Comparison of Floyd-Warshall Algorithm and Greedy Algorithm in Determining the Shortest Route,” \textit{2018 2nd East Indonesia Conference on Computer and Information Technology (EIConCIT)}, Makassar, Indonesia, 2018, pp. 294-298. doi:10.1109/eiconcit.2018.8878582  

\bibitem{laga} I. K. Laga Dwi Pandika, B. Irawan, and C. Setianingsih, “Apllication of Optimization Heavy Traffic Path with Floyd-Warshall Algorithm,” \textit{2018 International Conference on Control, Electronics, Renewable Energy and Communications (ICCEREC)}, Bandung, Indonesia, 2018, pp. 57-62. doi:10.1109/iccerec.2018.8712110 

\bibitem{alkan} M. ALKAN and M. AYDIN, “Simulation and Comparison of Pathfinding Algorithms using Real Turkey Data,” \textit{2018 International Conference on Artificial Intelligence and Data Processing (IDAP)}, Malatya, Turkey, 2018, pp. 1-4. doi:10.1109/idap.2018.8620849 

\bibitem{santis} R. De Santis, R. Montanari, G. Vignali, and E. Bottani, “An adapted ant colony optimization algorithm for the minimization of the travel distance of pickers in manual warehouses,” \textit{European Journal of Operational Research}, vol. 267, no. 1, pp. 120–137, 2018. doi:10.1016/j.ejor.2017.11.017

\bibitem{ning} J. Ning, Q. Zhang, C. Zhang, and B. Zhang, “A best-path-updating information-guided ant colony optimization algorithm,” \textit{Information Sciences}, vol. 433–434, pp. 142–162, 2018. doi:10.1016/j.ins.2017.12.047

\bibitem{pramudita} R. Pramudita \textit{et al.}, “Shortest Path Calculation Algorithms for Geographic Information Systems,” \textit{2019 Fourth International Conference on Informatics and Computing (ICIC)}, Semarang, Indonesia, 2019, pp. 1-5. doi:10.1109/icic47613.2019.8985871 

\bibitem{sangeetha} V. Sangeetha, R. Sivagami, R. Krishankumar, K. Ravichandran, and S. Tyagi, “A Modified Ant Colony Optimisation based Optimal Path finding on a Thematic Map,” \textit{2019 IEEE International Conference on Intelligent Techniques in Control, Optimization and Signal Processing (INCOS)}, Tamilnadu, India, 2019, pp. 1-5. doi:10.1109/incos45849.2019.8951373 

\bibitem{dermawan} T. S. Dermawan, “Comparison of Dijkstra dan Floyd-Warshall Algorithm to Determine the Best Route of Train," \textit{IJID (International Journal on Informatics for Development)}, vol. 7, no. 2, pp. 54–58, 2019.  doi:10.14421/ijid.2018.07202 

\bibitem{dondi} R. Dondi, G. Mauri, and I. Zoppis, “Graph algorithms,” \textit{Encyclopedia of Bioinformatics and Computational Biology}, vol. 1, pp. 940–949, 2019. doi:10.1016/b978-0-12-809633-8.20424-x 

\bibitem{aggarwal} S. Aggarwal and N. Kumar, “Path planning techniques for unmanned aerial vehicles: A review, solutions, and challenges,” \textit{Computer Communications}, vol. 149, pp. 270–299, 2020. doi:10.1016/j.comcom.2019.10.014 

\bibitem{daoqing} Z. Daoqing and J. Mingyan, “Parallel discrete lion swarm optimization algorithm for solving traveling salesman problem,” \textit{Journal of Systems Engineering and Electronics}, vol. 31, no. 4, pp. 751–760, 2020. doi:10.23919/jsee.2020.000050 

\bibitem{mesquita} R. Mesquita and P. D. Gaspar, “A Path Planning Optimization Algorithm Based on Particle Swarm Optimization for UAVs for Bird Monitoring and Repelling – Simulation Results,” \textit{2020 International Conference on Decision Aid Sciences and Application (DASA)}, Sakheer, Bahrain, 2020, pp. 1144-1148. doi:10.1109/dasa51403.2020.9317271 

\bibitem{arfananda} M. G. Arfananda, S. M. Nasution, and C. Setianingsih, “Selection of Bandung City Travel Route Using The FLOYD-WARSHALL Algorithm,” \textit{International Journal of Integrated Engineering}, vol. 12, no. 7, pp. 90-97, 2020. doi:10.30880/ijie.2020.12.07.010 

\bibitem{qu} C. Qu, W. Gai, J. Zhang, and M. Zhong, “A novel hybrid grey wolf optimizer algorithm for unmanned aerial vehicle (UAV) path planning,” \textit{Knowledge-Based Systems}, vol. 194, p. 105530, 2020. doi:10.1016/j.knosys.2020.105530 

\bibitem{gai} C. Qu, W. Gai, M. Zhong, and J. Zhang, “A novel reinforcement learning based grey wolf optimizer algorithm for unmanned aerial vehicles (uavs) path planning,” \textit{Applied Soft Computing}, vol. 89, p. 106099, 2020. doi:10.1016/j.asoc.2020.106099 

\bibitem{febryan} A. Z. Febryantika, F. A. Puspandini, I. R. Amalia, and M. Annisa, “Application of the Floyd Warshall algorithm in Determining the Shortest Route for Distribution of UD Nadira Cinta Rasa Bread to Praya, Central Lombok,” \textit{Eigen Mathematics Journal}, pp. 18–23, 2021. doi:10.29303/emj.v4i1.96 

\bibitem{hu} J. Zhang and A. Hu, “Analyzing green view index and green view index best path using Google street view and deep learning,” \textit{Journal of Computational Design and Engineering}, vol. 9, no. 5, pp. 2010–2023, 2022. doi:10.1093/jcde/qwac102 

\bibitem{sakharov} V. Sakharov, S. Chernyi, S. Saburov, and A. Chertkov, “Automatization search for the shortest routes in the transport network using the Floyd-warshell algorithm,” \textit{Transportation Research Procedia}, vol. 54, pp. 1–11, 2021. doi:10.1016/j.trpro.2021.02.041 

\bibitem{yuan} H. Yuan, J. Hu, Y. Song, Y. Li, and J. Du, “A new exact algorithm for the shortest path problem: An optimized shortest distance matrix,” \textit{Computers \& Industrial Engineering}, vol. 158, p. 107407, 2021. doi:10.1016/j.cie.2021.107407 

\bibitem{xie} X. Xie, Z. Tang, and J. Cai, “The multi-objective inspection path-planning in radioactive environment based on an improved ant colony optimization algorithm,” \textit{Progress in Nuclear Energy}, vol. 144, p. 104076, 2022. doi:10.1016/j.pnucene.2021.104076 

\bibitem{ntakolia} C. Ntakolia and D. V. Lyridis, “A comparative study on ant colony optimization algorithm approaches for solving multi-objective path planning problems in case of unmanned surface vehicles,” \textit{Ocean Engineering}, vol. 255, p. 111418, 2022. doi:10.1016/j.oceaneng.2022.111418 

\bibitem{yang} Z. Zhang and J. Yang, “A discrete cuckoo search algorithm for traveling salesman problem and its application in cutting path optimization,” \textit{Computers \& Industrial Engineering}, vol. 169, p. 108157, 2022. doi:10.1016/j.cie.2022.108157 

\bibitem{samosir} M. A. Samosir, “Application of the Dijkstra and Floyd–Warshall Algorithms in Determining the Shortest Route to Tourist Attractions in Toba,” \textit{Formosa Journal of Science and Technology}, vol. 2, no. 2, pp. 453–474, 2023. doi:10.55927/fjst.v2i2.2858 

\bibitem{wu} Z. Yan, J. Yan, Y. Wu, S. Cai, and H. Wang, “A novel reinforcement learning based tuna swarm optimization algorithm for autonomous underwater vehicle path planning,” \textit{Mathematics and Computers in Simulation}, vol. 209, pp. 55–86, 2023. doi:10.1016/j.matcom.2023.02.003

\bibitem{b42} Haddaway, N. R., Page, M. J., Pritchard, C. C., \& McGuinness, L. A. (2022). PRISMA2020: An R package and Shiny app for producing PRISMA 2020-compliant flow diagrams, with interactivity for optimised digital transparency and Open Synthesis Campbell Systematic Reviews, 18, e1230. https://doi.org/10.1002/cl2.1230

\end{thebibliography}
\end{document}